\documentclass[conference]{IEEEtran}
\IEEEoverridecommandlockouts

\usepackage{cite}
\usepackage{amsmath,amssymb,amsfonts}
\usepackage{graphicx}
\usepackage{svg}
\svgsetup{
  inkscapeexe=inkscape.com,
  inkscapeversion=1
}
\usepackage{textcomp}
\usepackage{xcolor}
\usepackage{booktabs}
\usepackage{url}

\def\BibTeX{{\rm B\kern-.05em{\sc i\kern-.025em b}\kern-.08em
    T\kern-.1667em\lower.7ex\hbox{E}\kern-.125emX}}

\begin{document}

\title{Towards Compact Sign Language Translation:\\
Frame Rate and Model Size Trade-offs
}

\author{
\IEEEauthorblockN{Kuanwei Chen\IEEEauthorrefmark{1}, Mengfeng Tsai}
\IEEEauthorblockA{Computer Science and Information Engineering, National Central University, Zhongli, Taiwan\\
elaping5691@gmail.com, mftsai@csie.ncu.edu.tw\\
\IEEEauthorrefmark{1}Corresponding author.}
}

\maketitle
\bstctlcite{T5SLTBSTcontrol}

\begin{abstract}
Sign Language Translation (SLT) converts sign language videos into
spoken-language text, bridging communication between Deaf and hearing
communities. Current gloss-free approaches rely on large encoder--decoder
models, limiting deployment. We propose a compact 77\,M-parameter
pipeline that couples MMPose skeletal pose extraction with a single linear
projection into T5-small. By varying the input frame rate we expose a
practical efficiency trade-off: at 12\,fps the model halves its sequence
length, achieving a 75\% reduction in encoder quadratic self-attention computational
complexity while incurring only a modest BLEU-4 drop (9.53 vs.\ 10.06 at 24\,fps on How2Sign).
Our system is roughly $3\times$ smaller than prior T5-base systems, demonstrating that a
lightweight architecture can remain competitive without hierarchical encoders
or large-scale models.
\end{abstract}


\section{Introduction}

Sign Language Translation (SLT) converts sign language video into text,
enabling accessibility for Deaf
communities~\cite{yin2021including}. Recent gloss-free methods pair
pretrained Transformer language models with video
encoders~\cite{camgoz2018neural}: Uthus et al.~\cite{uthus2023youtube}
used T5-base (${\sim}248$M parameters) with a linear projection, while
Rust et al.~\cite{rust2024privacy} added a hierarchical video encoder at
even greater cost. Pose-based methods reduce this overhead by representing
each frame as skeletal keypoints; Lin et al.~\cite{lin2023gloss} showed
that MMPose features with graph convolutions are competitive with
video-based encoders.

We propose a compact 77M-parameter architecture that uses
MMPose~\cite{mmpose2020} to extract 255-dimensional skeletal keypoints
and projects them into T5-small's embedding
space~\cite{raffel2020t5} via a single linear layer, eliminating
expensive video encoders. Our contributions are:
(1)~an efficient T5-small pipeline with a simple linear pose-to-token
projection;
(2)~a systematic analysis of the frame-rate/model-size/quality efficiency trade-off,
demonstrating that halving the frame rate reduces quadratic self-attention complexity by 75\%; and
(3)~evaluation on the How2Sign test set confirming the viability
of this lightweight design.

\begin{figure}[!t]
    \centering
    \includesvg[width=\columnwidth, inkscapelatex=false]{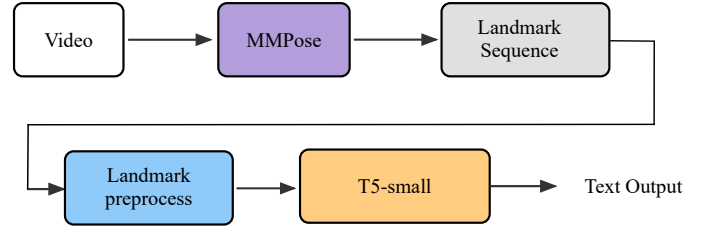}
    \caption{Proposed pipeline: MMPose extracts 255-dim pose features per frame,
        a linear layer maps them to 512-dim, and T5-small generates the translation.}
    \label{fig:arch}
\end{figure}

\section{Methodology}

\subsection{Preprocessing}
We use the SignDATA preprocessing pipeline~\cite{chen2026signdata} with
MMPose~\cite{mmpose2020} to detect 85 keypoints spanning both hands, the
upper body, and the face for every frame. Each keypoint contributes
$(x, y, z)$ coordinates, yielding a 255-dimensional vector per frame.
All coordinates are normalized by the frame width and height to achieve
scale invariance across video resolutions. Videos are resampled to a
uniform 24 fps to balance temporal coverage and computational cost;
we also evaluate at 12\,fps to study the speed--quality trade-off.

\subsection{Proposed Architecture}

The proposed pipeline is illustrated in Fig.~\ref{fig:arch}. A sign
language video is first processed by MMPose to produce a sequence of
255-dimensional pose vectors, one per frame. A single fully-connected
linear layer projects each vector from 255 to 512 dimensions, matching
T5-small's model dimension ($d_{\text{model}} = 512$).
The resulting sequence is fed into the T5-small encoder (6~Transformer layers),
and the decoder (6~layers) generates the English translation
autoregressively with beam search (beam size~5).
Per Table~1 of~\cite{raffel2020t5}, T5-small ($d_{\text{model}}{=}512$, $d_{\text{ff}}{=}2048$, 6~layers each,
${\approx}60$\,M) is $3.6\times$ smaller than T5-base ($d_{\text{model}}{=}768$, $d_{\text{ff}}{=}3072$, 12~layers each,
${\approx}220$\,M), with the halved depth and narrower width directly reducing computation.
The full model has approximately 77M parameters---roughly $3\times$
smaller than the T5-base systems employed by Uthus et al.~\cite{uthus2023youtube}---
with the projection layer adding negligible overhead.
We cap inputs at most 256 frames and limit the generated output to 128 tokens.

\subsection{Training}
Our model fine-tunes the pretrained \texttt{google/t5-v1\_1-small}
checkpoint without additional vision-encoder weights.
We use the AdaFactor optimizer at a constant learning rate of $1\times10^{-3}$ and no weight decay.
Training runs for 80 epochs with a 10-epoch linear warm-up.
We apply gradient clipping at norm~1.0 and label smoothing of~0.1, with effective batch size of~128.
The selected hyperparameters are summarized in Table~\ref{tab:hparams}.

\begin{table}[t]
\caption{Hyperparameter Choices for Training}
\label{tab:hparams}
\centering
\footnotesize
\resizebox{\columnwidth}{!}{%
\begin{tabular}{lll}
\toprule
\textbf{Hyperparameter} & \textbf{Candidate Values} & \textbf{Best (Ours)} \\
\midrule
Batch size & $\{32, 64, 128\}$ & $\mathbf{128}$ \\
Learning Rate (LR) & $\{1\mathrm{e}{-3}, 5\mathrm{e}{-4}, 1\mathrm{e}{-4}\}$ & $\mathbf{1\mathrm{e}{-3}}$ \\
Warm-up epochs & $\{5, 10, 20\}$ & $\mathbf{10}$ \\
Weight Decay & $\{0, 1\mathrm{e}{-3}, 1\mathrm{e}{-4}\}$ & $\mathbf{0}$ \\
Label Smoothing & $\{0.1, 0.2, 0.3\}$ & $\mathbf{0.1}$ \\
Dropout & $\{0, 0.1, 0.2, 0.3\}$ & $\mathbf{0.1}$ \\
\bottomrule
\end{tabular}
}
\end{table}

\section{Experiments}

\subsection{Datasets}
We use How2Sign~\cite{duarte2021how2sign} (originally {$\sim$}80\,h, 35k
clips) and YouTube-ASL~\cite{uthus2023youtube} (originally {$\sim$}984\,h, 610k captions).
All videos are resampled to 24\,fps; clips originally below this rate are discarded.
Each frame is then processed with MMPose into an 85-keypoint 3D pose sequence (255 values per frame).
We train on the combined corpora and evaluate on the How2Sign test set at both 24 and 12\,fps
(Section~\ref{sec:results}).

\subsection{Evaluation}
Translation quality is measured by BLEU-1 through BLEU-4, computed
with SacreBLEU~\cite{post2018call} under default tokenization.
Decoding uses beam size~5 with maximum output length~128.

\subsection{Results}\label{sec:results}
Table~\ref{tab:results} reports BLEU scores on the How2Sign test set
under three training configurations:
\begin{itemize}
\item \textbf{H2S}: trained on How2Sign only.
\item \textbf{YT-ASL}: trained on YouTube-ASL only.
\item \textbf{YT-ASL+H2S}: trained on a mixture of both datasets.
\end{itemize}
Joint training on YT-ASL+H2S consistently outperforms both single-dataset settings.
Increasing the frame rate from 12 to 24~fps improves BLEU-4 from 9.53 to 10.06,
indicating that finer temporal resolution benefits translation quality.
However, 12~fps halves the sequence length relative to 24~fps,
reducing the encoder's self-attention computational complexity by approximately 75\% due to its quadratic scaling ($O(n^2)$).
This presents a practical trade-off:
12~fps offers significantly faster training and inference with only a modest drop in performance,
while 24~fps maximizes translation quality.
Table~\ref{tab:results} also includes the T5-base results of Uthus et al.~\cite{uthus2023youtube} for reference.
While T5-base achieves a higher BLEU-4 of 11.89 on YT-ASL+H2S,
it uses ${\sim}248$M parameters---roughly $3\times$ larger than our 77M-parameter model.
Our best configuration (24\,fps, YT-ASL+H2S) reaches a BLEU-4 of 10.06,
demonstrating that a compact model can remain competitive with substantially larger systems.

\begin{table}[t]
    \caption{BLEU scores on the How2Sign test set.}
    \label{tab:results}
    \centering
    \footnotesize
    \setlength{\tabcolsep}{4pt}
    \begin{tabular}{@{}lcccc@{}}
        \toprule
        \textbf{Training Data} & \textbf{BLEU-1} & \textbf{BLEU-2} & \textbf{BLEU-3} & \textbf{BLEU-4} \\
        \midrule

        \multicolumn{5}{@{}l}{\textit{T5-base (248\,M params), 12-30\,fps~\cite{uthus2023youtube}}} \\
        H2S          & 14.96 &  5.11 &  2.26 &  1.22 \\
        YT-ASL       & 20.93 & 10.35 &  6.14 &  3.95 \\
        YT-ASL+H2S   & 36.35 & 23.00 & 16.13 & 11.89 \\
        \addlinespace

        \multicolumn{5}{@{}l}{\textit{Ours: T5-small (77\,M params), 12\,fps}} \\
        H2S          & 12.23 &  3.67 &  1.56 &  0.82 \\
        YT-ASL       & 17.30 &  7.82 &  5.28 &  3.34 \\
        YT-ASL+H2S   & 33.21 & 20.01 & 13.43 &  9.53 \\
        \addlinespace

        \multicolumn{5}{@{}l}{\textit{Ours: T5-small (77\,M params), 24\,fps}} \\
        H2S          & 12.98 &  4.22 &  1.82 &  1.07 \\
        YT-ASL       & 18.05 &  8.69 &  5.41 &  3.62 \\
        YT-ASL+H2S   & 34.29 & 20.83 & 14.06 & 10.06 \\
        \bottomrule
    \end{tabular}
\end{table}

\section{Conclusion}
We presented a compact 77\,M-parameter gloss-free ASL translation system pairing
MMPose pose extraction with a single linear projection into T5-small,
revealing clear trade-offs among model size, frame rate, and translation quality.
Our $3\times$ smaller model achieves BLEU-4 of 10.06 versus 11.89 for T5-base
(${\sim}248$M parameters), narrowing much of the gap with substantially fewer parameters.
At 12\,fps, halved sequence length reduces self-attention complexity by 75\%
while BLEU-4 drops only modestly to 9.53, confirming a practical path toward
efficient, real-time sign language translation.
Future work will explore frame-level temporal aggregation, larger T5 variants,
and data augmentation to further improvement.

\bibliographystyle{IEEEtran}
\bibliography{T5-SLT}

\end{document}